\documentclass[pre,floatfix,superscriptaddress,twocolumn,9pt]{revtex4-1}

\usepackage{amsmath,amssymb}
\usepackage{mathrsfs}
\usepackage{mathtools}
\usepackage{graphicx}% Include figure files
\usepackage{color}% Include colors for document elements
\usepackage{dcolumn}
\usepackage{hyperref}
\usepackage[normalem]{ulem}
\usepackage{algorithm}
\usepackage{extarrows}
\usepackage{subfig}
\usepackage{soul}
\usepackage{algpseudocode}
\usepackage{comment}

\algdef{SE}[DOWHILE]{Do}{doWhile}{\algorithmicdo}[1]{\algorithmicwhile\ #1}

\newcommand{\p}[1]{\left(#1\right)}

\newcommand{\abs}[1]{\left| #1 \right|} % for absolute value
\newcommand{\Expect}[1]{\mathop{\mathbb{E}}\left[#1\right]}
\newcommand{\Expectsub}[2]{\mathop{\mathbb{E}_{#1}}\left[#2\right]}

\newcommand{\norm}[1]{\left\lVert#1\right\rVert}

\newcommand{\myxlongequal}[1]{\stackrel{#1}{=}}
\usepackage[disable]{todonotes} 				% notes hidden
\newcommand*{\blauw}[1]{\textcolor{blue}{#1}}
\newcommand{\figurewidth}{.45\textwidth}

\begin{document}
	
	\title{Capabilities and Limitations of Time-lagged Autoencoders for Slow Mode Discovery in Dynamical Systems}
	
\author{Wei Chen}
\affiliation{
	Department of Physics, %
	University of Illinois at Urbana-Champaign, %
	1110 West Green Street, Urbana, Illinois 61801
}

\author{Hythem Sidky}
\affiliation{%
	Institute for Molecular Engineering, %
	5640 South Ellis Avenue, %
	University of Chicago, %
	Chicago, Illinois 60637%
}

\author{Andrew L. Ferguson}
\email{Author to whom correspondence should be addressed: \mbox{andrewferguson@uchicago.edu}}
\affiliation{%
	Institute for Molecular Engineering, %
	5640 South Ellis Avenue, %
	University of Chicago, %
	Chicago, Illinois 60637%
}
	
\begin{abstract}
	\noindent  Time-lagged autoencoders (TAEs) have been proposed as a deep learning regression-based approach to the discovery of slow modes in dynamical systems.  However, a rigorous analysis of nonlinear TAEs remains lacking.  In this work, we discuss the capabilities and limitations of TAEs through both theoretical and numerical analyses.  Theoretically, we derive bounds for nonlinear TAE performance in slow mode discovery and show that in general TAEs learn a mixture of slow and maximum variance modes. Numerically, we illustrate cases where TAEs can and cannot correctly identify the leading slowest mode in two example systems: a 2D ``Washington beltway'' potential and the alanine dipeptide molecule in explicit water.  We also compare the TAE results with those obtained using state-free reversible VAMPnets (SRVs) as a variational-based neural network approach for slow modes discovery, and show that SRVs can correctly discover slow modes where TAEs fail.
\end{abstract}

\maketitle

	% \begin{wileykeywords}
	% tICA, kernel tICA, hierarchical dynamics encoder, slow modes learning, neural network, deep learning
	% \end{wileykeywords}

	%\section*{\sffamily \Large First-order heading}
	%\subsection*{\sffamily \large Second-order heading}
	%\subsubsection*{\sffamily \normalsize Third-order heading}
	%{\sffamily \small Fourth-order heading}\\

\section*{Introduction}

Estimation of the slow (i.e., maximally autocorrelated) collective modes of a dynamical system from trajectory data is an important topic in dynamical systems theory in understanding, predicting, and controlling long-time system evolution \cite{andrew2013deep, mardt2018vampnets, pathak2018model, Ye_2015, giannakis2012nonlinear,korda2018convergence,sharma2016correspondence,korda2018linear,koopman1931hamiltonian,mezic2005spectral}. In the context of molecular dynamics, identification of the leading slow modes is of great value in illuminating conformational mechanisms, constructing long-time kinetic models, and guiding enhanced sampling techniques \cite{noe2013variational, nuske2014variational, schutte2001transfer, prinz2011markov, schwantes2015modeling, schwantes2013improvements, pande2010everything, chodera2014markov, chodera2014markov, perez2013identification, schwantes2015modeling, harrigan2017landmark, mardt2018vampnets, hernandez2018variational, sultan2018transferable, wehmeyer2018time}. Many machine learning models have been applied to learn slow modes from molecular trajectory data, some falling into the category of more traditional techniques, including time-lagged independent component analysis (TICA) \cite{perez2013identification, noe2013variational, nuske2014variational, noe2015kinetic, noe2016commute,	perez2016hierarchical, schwantes2013improvements, klus2018data,husic2018markov}, kernel TICA \cite{schwantes2015modeling,harrigan2017landmark}, Markov state models (MSMs) \cite{prinz2011markov,pande2010everything,schwantes2013improvements,husic2018markov,trendelkamp2015estimation,sultan2017transfer,mittal2018recruiting,harrigan2017msmbuilder,wehmeyer2018introduction,scherer2018variational}, while others employ more recently-developed deep learning models, including time-lagged autoencoders (TAEs) \cite{wehmeyer2018time}, variational dynamics encoders (VDEs) \cite{wayment2018note,hernandez2017variational,sultan2018transferable}, variational approach for Markov processes nets (VAMPnets) \cite{mardt2018vampnets}, and state-free reversible VAMPnets (SRVs) \cite{chen2019nonlinear}. These approaches all employ variants of deep neural networks, but differ in the details of their architecture and loss function: TAEs are a regression approach that minimize time-lagged reconstruction loss, VAMPnets and SRVs are variational approaches that maximize autocorrelation of the slow modes, and VDEs can be conceived as a mixture of the regression and variational approaches.  Although there are existing theoretical guarantees of slow modes discovery for variational approaches \cite{chen2019nonlinear,noe2013variational,nuske2014variational, schutte2001transfer}, similar theoretical guarantees for regression approaches is currently limited to linear cases \cite{wehmeyer2018time}. Specifically, linear TAEs are known to be equivalent to time-lagged canonical correlation analysis, and closely related to  TICA and kinetic maps \cite{wehmeyer2018time, wu2017variational, hotelling1936relations, noe2015kinetic, perez2013identification}.  In this work, we aim to fill this gap by presenting a theoretical and numerical analysis of the capabilities and limitations of TAEs as a nonlinear regression approach for slow mode discovery. 

\section*{Results and Discussion}

\begin{figure*}[ht!]
\begin{center}
	\includegraphics[width=0.59\textwidth]{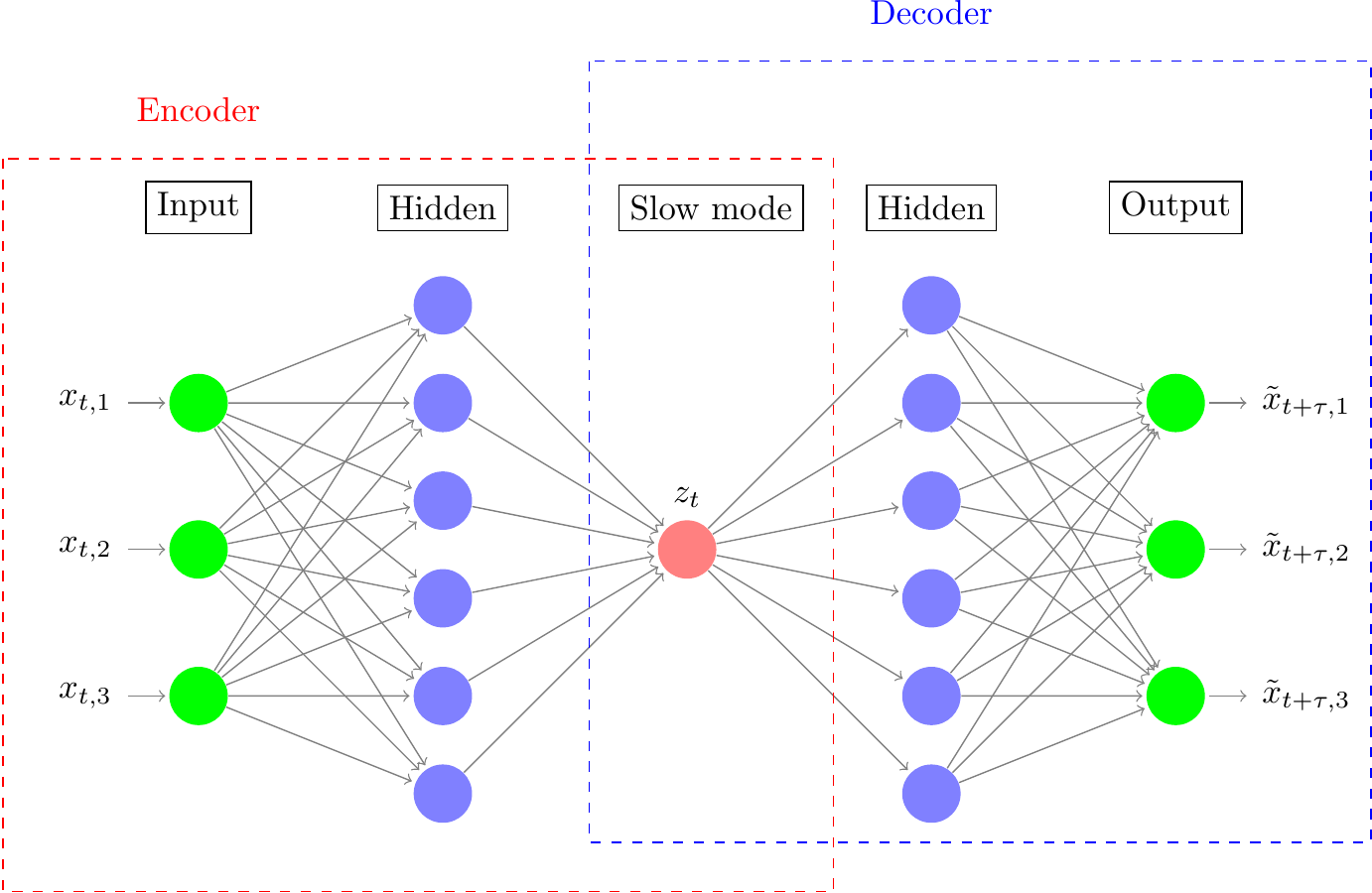}
	\caption{Schematic of a time-lagged autoencoder (TAE). The input configuration $x_t=(x_{t,1},x_{t,2},x_{t,3})$ at time $t$ is fed into an encoder network to generate the latent space encoding $z_t=E(x_t)$. This encoding is passed into a decoder network to generate output $\tilde{x}_{t+\tau}=(\tilde{x}_{t+\tau,1},\tilde{x}_{t+\tau,2},\tilde{x}_{t+\tau,3})=D(z_t)$ which aims to reconstruct the configuration $x_{t+\tau}$ at a later time $t+\tau$.  Training is performed by backpropagation and is terminated when the loss $d_\tau=\Expect{\norm{\tilde{x}_{t+\tau}-x_{t+\tau}}^2}$ is minimized.  Image constructed using code downloaded from \url{http://www.texample.net/tikz/examples/neural-network} with permission of the author Kjell Magne Fauske.}
	\label{TAE_diagram}
\end{center}
\end{figure*}

Consider a trajectory of a dynamical system $\{x_t\}$ where $x_t$ is a system configuration, or a derived featurization of the configuration, at time $t$. We define the slowest mode for a given lag time $\tau$ as the functional mapping $z(\cdot)$ that maximizes autocorrelation $A(z)$ for a lag time $\tau$,
\begin{equation}\label{slow_mode_def}
A(z) = \frac{\Expect{\delta z(x_t)\delta z(x_{t+\tau})}}{\sigma^2(z)},
\end{equation} 
where $\delta z(x_t)=z(x_t)-\Expect{z(x_t)}$ is the mean-free slow mode and $\sigma^2(z)$ is its variance.  This definition is closely related to the dominant eigenfunction of the transfer operator \cite{chen2019nonlinear,noe2013variational,nuske2014variational, schutte2001transfer}. A TAE seeks to estimate  the  slowest mode by training a time-lagged autoencoder to  encode a system configuration $x_t$ at time $t$ into a low-dimensional latent  space $z_t=E(x_t)$, and then decode this latent space embedding to reconstruct the system configuration $x_{t+\tau} =D(z_t) = D(E(x_t))$ at time $(t+\tau)$. The operational principle is that minimizing the reconstruction loss $\Expect{\norm{D(E(x_t))-x_{t+\tau}}^2}$ at a lag time $\tau$ promotes discovery of slow modes $z_t=E(x_t)$ within  the latent space. In the following sections, we define under what conditions TAEs are able to correctly learn the slowest mode and when they will fail to do so.  To simplify our discussion, we restrict our analysis to recovery  of the leading slowest mode of the system. A schematic of a fully-connected feedforward TAE with a 1D latent  space is presented in \blauw{Fig.~\ref{TAE_diagram}}.

\subsection{Linear time-lagged autoencoders (TAEs) can learn the slowest mode by  employing whitened features}

Consider a sufficiently long 2D trajectory $\{x_t\}=\{(x_{t,1},x_{t,2})\}$ for a stationary process such that the mean and variance do not change over time and the two components $x_{t,1}$ and $x_{t,2}$ are mean-free and mutually independent.  Let the autocorrelation and the variance for component $i\ (i=1,2)$ be $A(x_{t,i})$ and $\sigma^2(x_{t,i})$, respectively.  Let $z_t=E(x_t)$ be the latent variable, where $E$ is the encoder mapping for TAE, and $\tilde{x}_{t+\tau}=D(z_t)$ be the reconstructed time-lagged output, where $D$ is the decoder.  The TAE seeks to find the encoding and decoding functional mappings $E$ and $D$ to minimize the time-lagged reconstruction loss,
\begin{align}\label{loss}
d_\tau&=\Expect{\norm{\tilde{x}_{t+\tau}-x_{t+\tau}}^2}\nonumber\\
&=\Expect{\norm{D(z_t)-x_{t+\tau}}^2}\nonumber\\
&=\Expect{\norm{D(E(x_t))-x_{t+\tau}}^2}.
\end{align}
Let us assume that $A(x_{t,1}) > A(x_{t,2})$, which defines $x_{t,1}$ to be a slower component than $x_{t,2}$.  If $z\sim x_{t,1}$ (denoting that $z$ is a bijection of $x_{t,1}$, which in the linear case is a  non-trivial linear transformation), the reconstructed output $D(z)$ should be a linear function of $x_{t,1}$ given by,
\begin{equation}\label{decoder_linear}
D(z)=(c_1x_{t,1}+c_0, 0)\ (c_1\neq 0),
\end{equation}
where $c_1$ and $c_0$ are constants, and the second component is 0 since $z$ does not contain information about $x_{t,2}$.   The corresponding time-lagged reconstruction loss is given by,
\begin{widetext}
\begin{align}\label{tae_loss_linear}
d_\tau(z\sim x_{t,1})
&=\Expect{\norm{D(z)-x_{t+\tau}}^2}\nonumber\\
&=\Expect{\norm{(c_1x_{t,1}+c_0, 0)-(x_{t+\tau,1},x_{t+\tau,2})}^2}\nonumber\\
&\myxlongequal{\text{mean-free features}}\Expect{c_1^2x_{t, 1}^2+c_0^2+x_{t+\tau, 1}^2-2x_{t+\tau,1}c_1x_{t,1} + x_{t+\tau,2}^2}\nonumber\\
&\myxlongequal{\text{process is stationary}}\Expect{c_1^2x_{t, 1}^2+c_0^2+x_{t, 1}^2-2x_{t+\tau,1}c_1x_{t,1} + x_{t,2}^2}.\nonumber\\
\end{align}
\end{widetext}
The optimal reconstruction coefficients are given by minimization of \blauw{Eq.~\ref{tae_loss_linear}} with respect to $c_0$ and $c_1$,
\begin{equation}\label{opt_coeff}
\begin{cases}
c_0&=0,\\
c_1&=\Expect{x_{t,1}x_{t+\tau,1}}/\Expect{x_{t,1}^2}=A(x_{t,1}),
\end{cases}
\end{equation}
and the corresponding minimal loss is given by,
\begin{equation}\label{tae_loss_linear_1}
d_\tau(z\sim x_{t,1}) = \sigma^2(x_{t,1})(1-A^2(x_{t,1}))+\sigma^2(x_{t,2}),
\end{equation} 
where we employed the substitution $\Expect{x_{t,1}x_{t+\tau,1}}=A(x_{t,1})\Expect{x_{t,1}^2}=A(x_{t,1})\sigma^2({x_{t,1}^2})$. We identify the first term $\sigma^2(x_{t,1})(1-A^2(x_{t,1}))$ as the ``propagation loss'', which increases as autocorrelation decreases and therefore generally increases with lag time $\tau$.  We identify the second term $\sigma^2(x_{t,2})$ as the ``irreducible capacity loss'', which reflects the fact that the one-dimensional latent variable $z$ does not contain any information about component $x_{t,2}$, and is independent of lag time.  
\blauw{Eq.~\ref{tae_loss_linear_1}} can be rearranged as,
\begin{align}\label{tl_loss_1}
&d_\tau(z\sim x_{t,1})\nonumber\\
=&\p{\sigma^2(x_{t,1})+\sigma^2(x_{t,2})}-\sigma^2(x_{t,1})A^2(x_{t,1})\nonumber\\
=&\sigma^2(x)-\sigma^2(x_{t,1})A^2(x_{t,1}),
\end{align}
where $\sigma^2(x)$ is the time-independent total variance of configurations.

If we now consider the case that $z\sim x_{t,2}$ by an analogous analysis the loss is given by,
\begin{align}\label{tl_loss_2}
&d_\tau(z\sim x_{t,2})\nonumber\\
=&\sigma^2(x)-\sigma^2(x_{t,2})A^2(x_{t,2})
\end{align}

From \blauw{Eq.~\ref{tl_loss_1}} and \blauw{Eq.~\ref{tl_loss_2}}, we see that in the time-lagged reconstruction loss there are two contributing factors: variance and autocorrelation. By construction $A(x_{t,1}) > A(x_{t,2})$, so it is the objective of the TAE to learn $z\sim x_{t,1}$ as the slowest mode. However, if $\sigma^2(x_{t,2})$ is sufficiently large compared to $\sigma^2(x_{t,1})$ such that,
\begin{equation}\label{key}
\sigma^2(x_{t,1})A^2(x_{t,1})<\sigma^2(x_{t,2})A^2(x_{t,2}),
\end{equation}
then,
\begin{equation}\label{key1}
d_\tau(z\sim x_{t,1}) > d_\tau(z\sim x_{t,2}),
\end{equation}
and the TAE loss is minimized by learning $z\sim x_{t,2}$.

%In this case, suppose the output of the decoder is
%\begin{equation}\label{key2}
%D(z)=(c_1\p{b_1x_{t,1}+b_2x_{t,2}}+c_{10}, c_2\p{b_1x_{t,1}+b_2x_{t,2}}+c_{20})
%\end{equation}

%\weiadd{Then the loss function is }
%\begin{align}\label{key3}
%&d_\tau(z\sim b_1x_{t,1}+b_2x_{t,2})\nonumber\\
%=&\Expect{\norm{D(z)-x_{t+\tau}}^2}\nonumber\\
%=&\p{c_1\p{b_1x_{t,1}+b_2x_{t,2}}+c_{10}-x_{t+\tau,1}}^2\nonumber\\
%+&\p{c_2\p{b_1x_{t,1}+b_2x_{t,2}}+c_{20}-x_{t+\tau,2}}^2
%\end{align}

%\weiadd{The optimal coefficients are}
%\begin{equation}\label{key4}
%\begin{cases}
%c_1&=\frac{b_1\sigma_1^2A_1}{b_1^2\sigma_1^2+b_2^2\sigma_2^2}\nonumber\\
%c_2&=\frac{b_2\sigma_2^2A_2}{b_1^2\sigma_1^2+b_2^2\sigma_2^2}\nonumber\\
%c_{10}&=c_{20}=0\nonumber\\
%A_i&=A(x_{t,i})\ (i=1,2)\nonumber\\
%\sigma_i&=\sigma(x_{t,i})\ (i=1,2)
%\end{cases}
%\end{equation}

We also consider whether it is possible that the optimal linear mode is actually a mixed linear mode where $z\sim (b_1x_{t,1}+b_2x_{t,2})$ and $b_1^2+b_2^2=1$.  It can be shown by an analogous analysis, and recalling that the two components are independent and mean free, that the minimal loss is,
%\begin{align}\label{key5}
%&d_\tau(z\sim b_1x_{t,1}+b_2x_{t,2})\nonumber\\
%=&\p{\sigma_1^2+\sigma_2^2}-\frac{b_1^2\sigma_1^4A_1^2+b_2^2\sigma_2^4A_2^2}%{b_1^2\sigma_1^2+b_2^2\sigma_2^2},
%\end{align}
\begin{align}\label{key6}
&d_\tau(z\sim (b_1x_{t,1}+b_2x_{t,2}))\nonumber\\
=&\sigma^2(x)-\frac{b_1^2\sigma^4(x_{t,1})A^2(x_{t,1})+b_2^2\sigma^4(x_{t,2})A^2(x_{t,2})}{b_1^2\sigma^2(x_{t,1})+b_2^2\sigma^2(x_{t,2})},
\end{align}
and that this expression reduces to \blauw{Eq.~\ref{tl_loss_1}} for $b_2=0$ and \blauw{Eq.~\ref{tl_loss_2}} for $b_1=0$ as expected.

Using $b_1^2+b_2^2=1$ to eliminate $b_1$, the minimal loss can be simplified to,
%\begin{align}\label{key7}
%&d_\tau(z\sim b_1x_{t,1}+b_2x_{t,2})\nonumber\\
%=&\p{\sigma_1^2+\sigma_2^2}-\sigma_1^2A_1^2-\frac{\sigma_2^2(\sigma_2^2A_2^2-\sigma_1^2A_1^2)}{(1/b_2^2-1)\sigma_1^2+\sigma_2^2}
%\end{align}
\begin{widetext}
\begin{align}\label{key8}
d_\tau(z\sim b_1x_{t,1}+b_2x_{t,2}) =
\begin{cases}
        \sigma^2(x)-\sigma^2(x_{t,1})A^2(x_{t,1}), & \text{for } b_2 = 0\\
        \sigma^2(x)-\sigma^2(x_{t,1})A^2(x_{t,1})-\frac{\sigma^2(x_{t,2})(\sigma^2(x_{t,2})A^2(x_{t,2})-\sigma^2(x_{t,1})A^2(x_{t,1}))}{(1/b_2^2-1)\sigma^2(x_{t,1})+\sigma^2(x_{t,2})}, & \text{for } 0< b_2 < 1\\
        \sigma^2(x)-\sigma^2(x_{t,2})A^2(x_{t,2}), & \text{for } b_2 = 1
        \end{cases}
\end{align}
\end{widetext}
which is a monotonic function with respect to $b_2$. Recalling that all variances and squared autocorrelations constrained to non-negative values, if $\sigma^2(x_{t,1})A^2(x_{t,1}) < \sigma^2(x_{t,2})A^2(x_{t,2})$ then the loss is minimized for $b_2=1$ and $z\sim x_{t,2}$, whereas if $\sigma^2(x_{t,1})A^2(x_{t,1}) > \sigma^2(x_{t,2})A^2(x_{t,2})$ then the loss is minimized for $b_2=0$ and $z\sim x_{t,1}$. Accordingly, except for the case $\sigma^2(x_{t,1})A^2(x_{t,1}) = \sigma^2(x_{t,2})A^2(x_{t,2})$, the loss is globally minimized by learning one of the pure component modes, the optimum is not a mixture of the two modes, and the analysis presented for the two pure modes is sufficient and complete for the case of a linear TAE with independent components.

This theoretical development demonstrates that a linear TAE is not guaranteed to find the slowest mode in the event that the associated variance of a faster mode is sufficiently large such  that its erroneous identification leads to a smaller reconstruction loss. A straightforward solution to this issue is to apply a whitening transformation \cite{wehmeyer2018time} to the input data such that $\sigma^2(x_{t,1})=\sigma^2(x_{t,2})=1$, and any linear combination $b_1 x_{t,1}+b_2 x_{t,2}$ with $b_1^2+b_2^2=1$ has unit variance. By eliminating the variance of the learned mode as a discriminating feature within the loss functions \blauw{Eq.~\ref{tl_loss_1}} and \blauw{Eq.~\ref{tl_loss_2}}, learning is performed exclusively on the basis of autocorrelation, and the linear TAE can correctly learn the slowest mode by minimizing the reconstruction loss. A rigorous proof that linear TAEs employing whitened features is equivalent to TICA for reversible process and can correctly identify the slowest mode is presented in Ref.\ \cite{wehmeyer2018time}.

\subsection{Nonlinear TAEs cannot equalize the variance explained within the input features and so cannot be assured to learn the slowest mode}   

We now proceed to perform a similar analysis for nonlinear TAEs.  To aid in our discussion, we define $\bar{x}(z)=\Expectsub{x}{x|z}$ and introduce the ``variance explained'' $\sigma^2(\bar{x}(z))$ by the latent variable $z$ as, 
\begin{equation}\label{ve}
\sigma^2(\bar{x}(z))=\sigma^2(\Expectsub{x}{x|z})
\end{equation}
which measures how much variance in the feature space can be explained by $z$.  
The idea of this concept is as follows. In a standard (non-time-lagged) autoencoder, the optimal reconstruction $\tilde{x}=D(z)$ is $D(z)=\Expectsub{x}{x|z}$ and the variance of the outputs is $\sigma^2(\Expectsub{x}{x|z})$, which explains part of the variance in the feature space while leaving $\p{\sigma^2(x)-\sigma^2(\Expectsub{x}{x|z})}$ unexplained.  In analogy to the linear case, it may be shown that for a long trajectory generated by a stationary process with finite states, the optimal loss for nonlinear TAE is approximately 
\begin{equation}\label{lower_bound_1}
d_\tau(z)\approx  \sigma^2(\bar{x}(z))(1-G^2(z))+\p{\sigma^2(x)-\sigma^2(\bar{x}(z))}
\end{equation}
where $G(z)$ is the nonlinear generalization of autocorrelation $A(z)$ and $\sigma^2(x)$ is the total variance of the input features.  Here $\sigma^2(\bar{x}(z))(1-G^2(z))$ is the ``propagation loss'' and $\p{\sigma^2(x)-\sigma^2(\bar{x}(z))}$ is the ``irreducible capacity loss'' for the nonlinear case. The details of the proof and related concepts can be found in Section \ref{theory_bounds} of the \blauw{Appendix}.

From \blauw{Eq.~\ref{lower_bound_1}}, we see that both variance explained and the autocorrelation contribute to the TAE loss as in the linear case.  If a faster mode has much larger variance explained than the true slowest mode, it is possible that nonlinear TAE would learn this faster mode in the latent encoding to minimize the loss, and in this case the nonlinear TAE fails to learn the correct slowest mode.

\begin{figure}[ht!]
\begin{center}
	\includegraphics[width=0.40\textwidth]{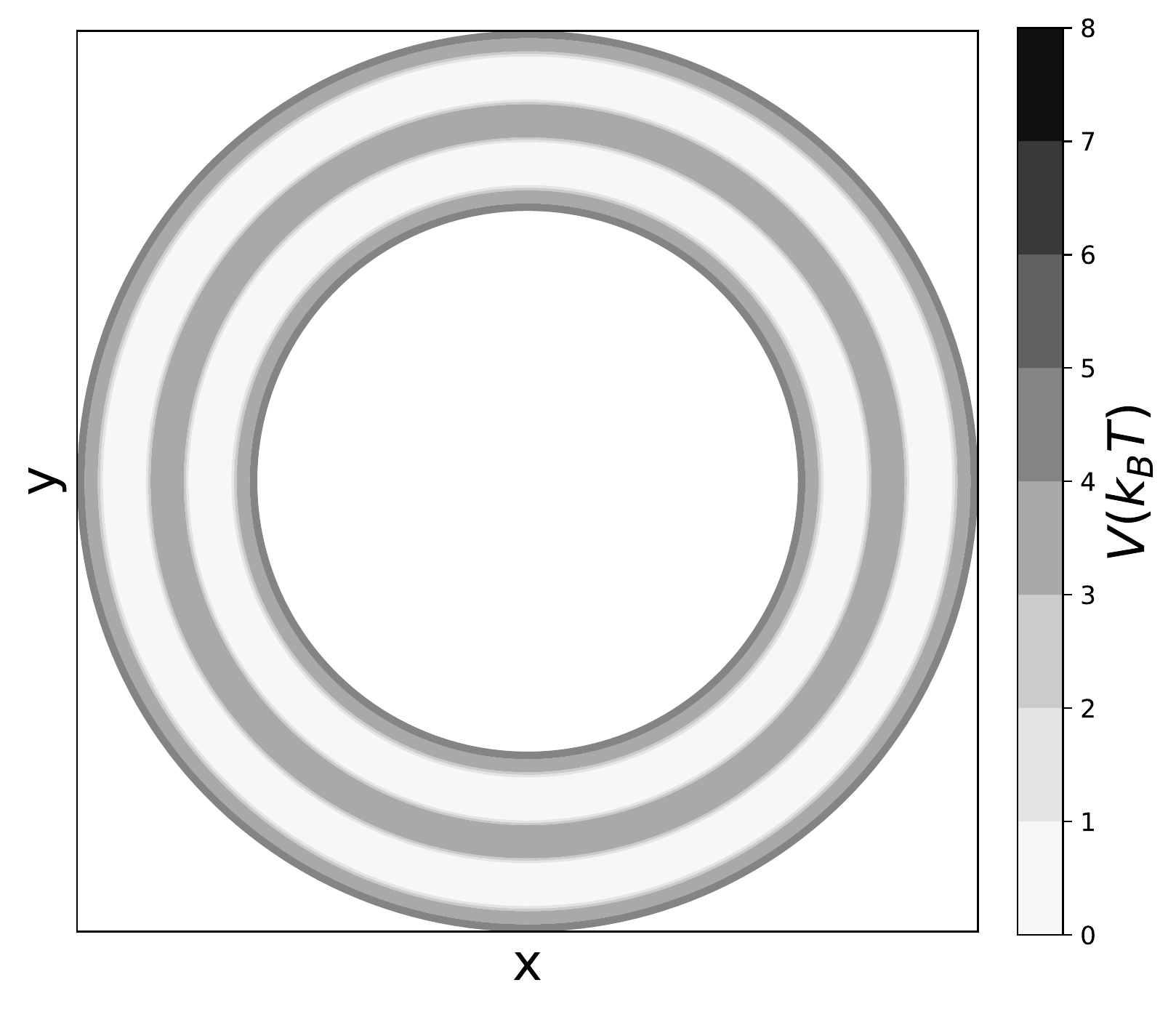}
	\caption{Contour plot of the Washington beltway potential, which consists two circular potential valleys at 0 $k_B T$ separated by a circular barrier of 4 $k_BT$. }
	\label{C_potential}
\end{center}
\end{figure}

We demonstrate this idea in the context of ``Washington beltway'' potential, which consists two circular potential valleys at 0 $k_BT$, separated by a circular barrier of 4 $k_BT$ (\blauw{Fig.~\ref{C_potential}}).  The expression for the potential is given by 
\begin{equation}\label{potential_C_exp}
\frac{V(r, \theta)}{k_BT} = \begin{cases}
0\text{,  if } r_1-dr < r < r_1+dr \\
0\text{,  if } r_2-dr < r < r_2+dr \\
4\text{,  if } r_1+dr < r < r_2-dr \\

1.25 + 7.5 \p{\abs{r - r_1}  + \abs{r - r_2}} \text{, otherwise}
\end{cases}
\end{equation}
where $r_1=0.7, r_2=0.9, dr=0.05$.  To simulate a particle moving in this potential, we conduct a Markov state model (MSM) simulation to generate a trajectory following the procedure detailed in Ref.~\cite{chen2019nonlinear}.  Specifically, we split $(r,\theta)\in [0.6,1]\times[0,2\pi]$ into 20-by-200 evenly spaced bins, and run a 5,000,000 step MSM simulation according to transition probabilities from bin $i$ to bin $j$,
\begin{equation}\label{transition_prop_C}
p_{ij} = \begin{cases}
C_i e^{-(V_j-V_i) / (k_BT)}, \ \text{if $i$,$j$ are neighbors or $i$=$j$}\\
0, \ \text{otherwise}
\end{cases}
\end{equation}
where $C_i$ is the normalization factor that ensures $\sum_j p_{ij} = 1$. 

\begin{figure*}[ht!]
	\begin{center}
		\includegraphics[width=0.80\textwidth]{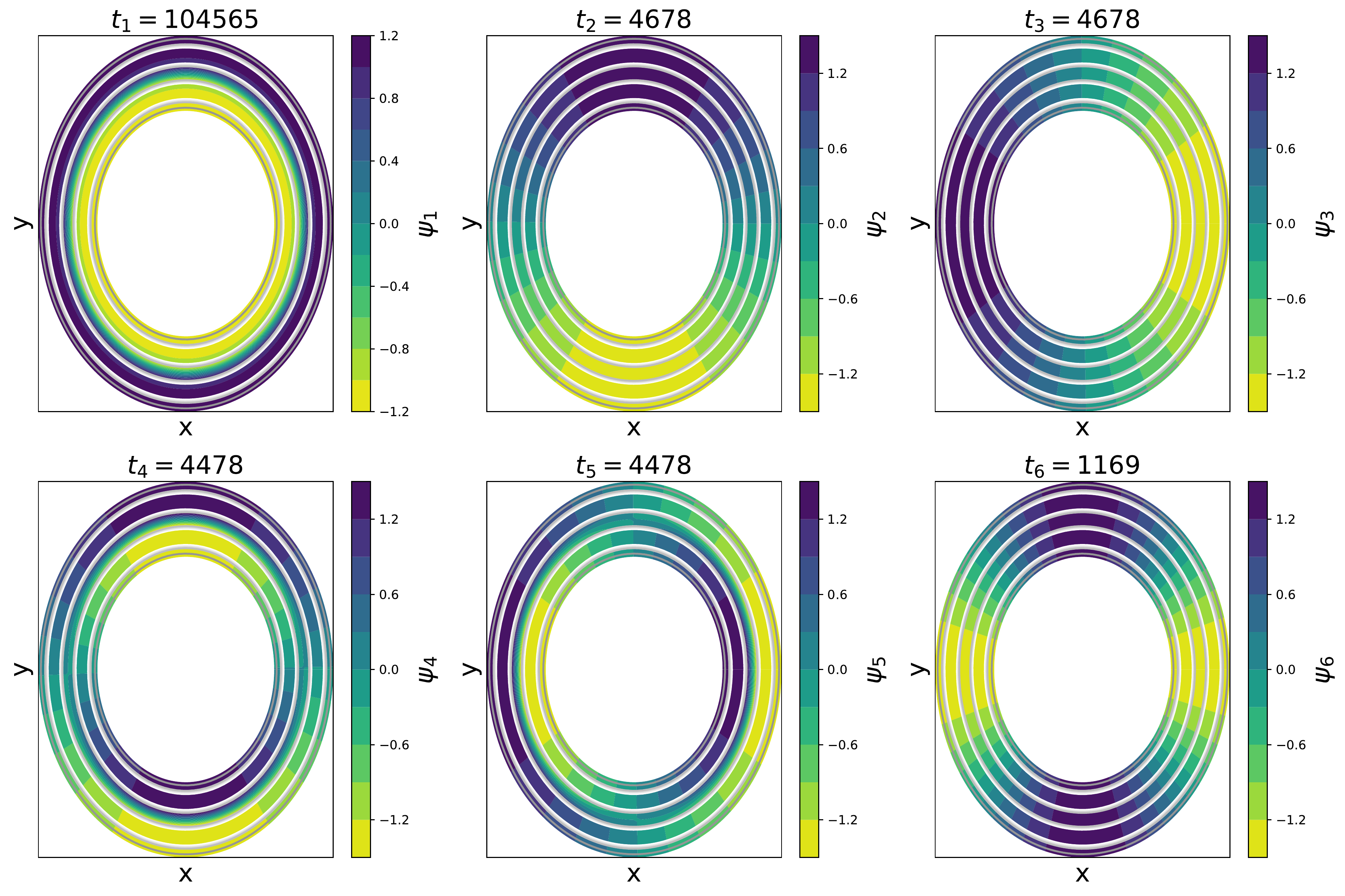}
		\caption{First six leading slow modes of the Washington beltway potential with corresponding timescales marked above each subplot.  The slowest mode ($t_1$ = 104,565) corresponds to transitions in the $r$ direction over the circular barrier separating the two circular valleys. The next two slowest modes form a degenerate pair ($t_2$ = $t_3$ = 4678) corresponding to transitions around the circular valleys in the $\theta$ direction. The higher order modes correspond to mixed $r$-$\theta$ transitions and appear as degenerate pairs due to the circular symmetry in $\theta$. }
		\label{C_potential_slow_modes}
	\end{center}
\end{figure*}

Given the analytical expression for the potential, the slow modes are analytically calculable by computing eigenvectors of the MSM transition matrix.  The first six leading slow modes with their implied timescales are presented in \blauw{Fig.~\ref{C_potential_slow_modes}}.  As expected, the slowest mode corresponds to transitions over the potential barrier radial direction ($r$-direction). Transitions around the circular potential in the polar direction ($\theta$-direction) appear as a degenerate pair within the second and third slowest modes with an order of magnitude faster timescale. Higher-order modes correspond to mixed $r$-$\theta$ transitions and appear as degenerate pairs due to the circular symmetry in $\theta$.

\begin{figure*}[ht!]
	\begin{center}
		\includegraphics[width=0.58\textwidth]{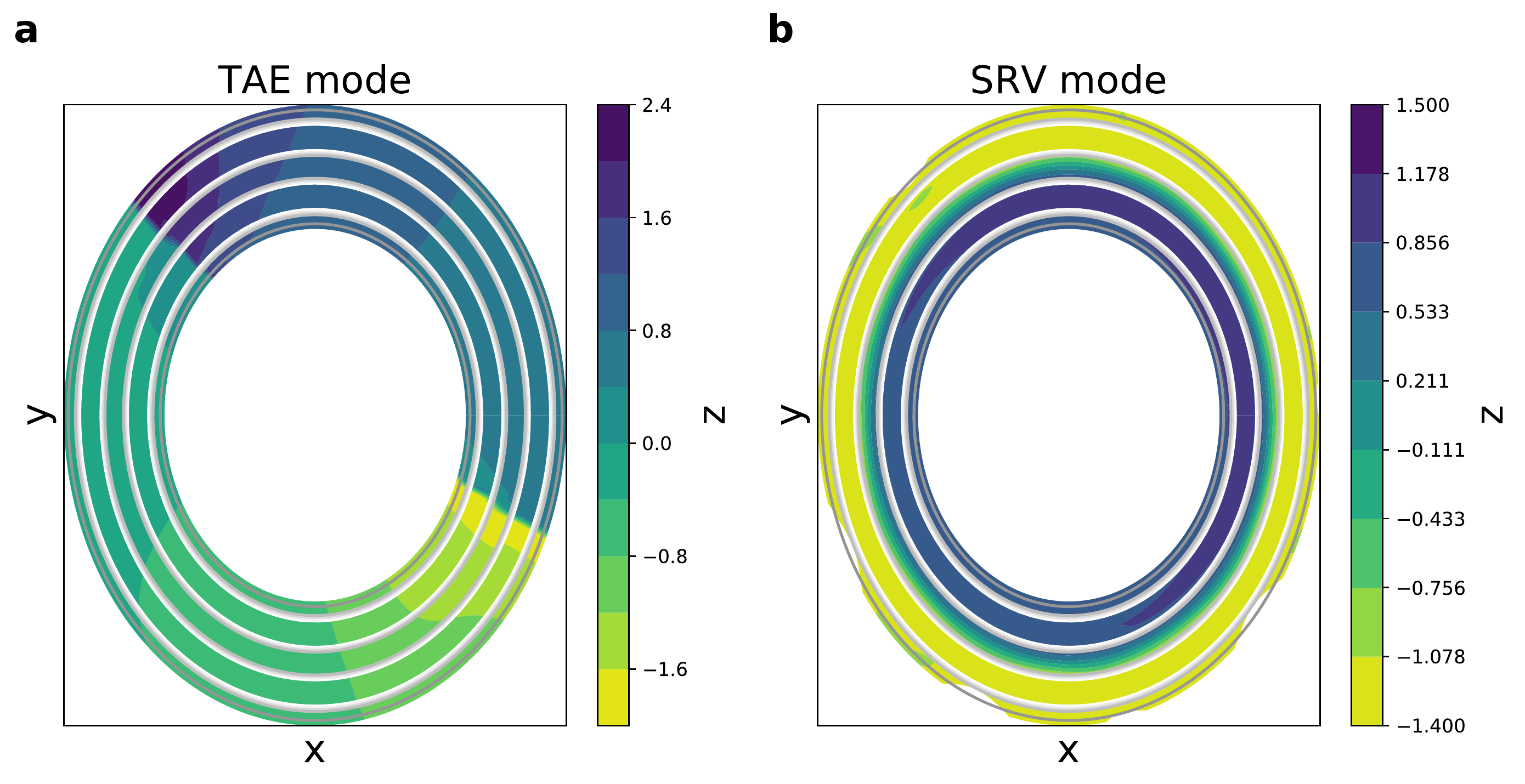}
		\caption{Slowest mode discovered by TAE and SRV. (a) The TAE incorrectly identifies $\theta$ as the slowest mode since the $\theta$ direction has much larger variance explained and therefore much larger contribution to the TAE loss. (b) The SRV successfully discovers $r$ direction as the slowest mode. }
		\label{C_TAE_CV}
	\end{center}
\end{figure*}

We now analyze the MSM trajectory using a nonlinear TAE with a [2-50-50-1-50-50-2] architecture and tanh activation functions for encoding and decoding layers and linear activation functions for input/output/latent layers. We supply to the input and output layers $(x(t),y(t))$ and $(x(t+\tau),y(t+\tau))$ pairs, respectively, employing a lag time of $\tau$ = 3000 steps. The data are naturally mean-free due to the topology of the potential landscape and are pre-whitened to normalize their variance. In \blauw{Fig.~\ref{C_TAE_CV}a}, we show the contour plot of the latent variable for the system.  Clearly, the TAE did not correctly identify theoretical slowest mode, which should be along $r$ direction. Instead, since $\theta$-direction has much larger variance explained than $r$-direction, the TAE is biased towards identifying $\theta$ as slower mode in order to minimize time-lagged reconstruction loss.  Therefore the learned latent variable is actually the mixture of $r$ and $\theta$ with most contribution coming from $\theta$. 

To be quantitative, we compare the TAE loss employing $r$ and $\theta$ as the latent variable. Due to symmetry, when $z\sim r$ (denoting that $z$ is a bijection of $r$), the average configuration given $r$ is $\bar{x}(z\sim r)=0$, so $\sigma^2(\bar{x}(z\sim r))=0$ and $d_\tau(z\sim r)=\sigma^2(x)$ from \blauw{Eq.~\ref{lower_bound_1}}. Conversely if $z\sim \theta$, the average configuration given $\theta$ is $\bar{x}(z\sim \theta)=(\Expect{r}\cos\theta, \Expect{r}\sin\theta)$, so $\sigma^2(\bar{x}(z\sim \theta))=\Expect{r}^2$ and $d_\tau(z\sim \theta)=\sigma^2(x)-\Expect{r}^2G^2(z\sim\theta)$.  For $\Expect{r}^2G^2(z\sim\theta)>0$, learning $z\sim\theta$ as the slow mode results in a lower time-lagged reconstruction loss than learning the true slow mode $z\sim r$. We numerically estimate from the simulation trajectory $G(z\sim \theta)\approx 0.535, \sigma^2(x)=2$, and $\Expect{r}\approx 1.4$, from which we compute $d_\tau(z\sim r) = 2$  and then the estimated loss $d_\tau(z\sim\theta) \approx 1.44$. The actual training loss is computed to be $d_\tau \approx 1.48$, showing that the nonlinear TAE approximately learns $\theta$ as the slowest mode. As a side note, if we use an optimal encoding corresponding to a bijective encoding of the feature space (see last part of Section \ref{theory_bounds} of the \blauw{Appendix} for details), the minimal possible loss is $d_\tau \approx 1.42$, which implies that $\theta$ is very close to the optimal encoding. 

%For alanine dipeptide, it may not be good to do bound estimation, since the theorem only strictly works for finite-state models.  For infinite-state model like alanine dipeptide, the problem is: given any countable length of trajectory, the probability of having two states with equal non-trivial $z$ (e.g. $\phi$ or $\psi$) is 0.  Therefore in theory, a large enough network (without regularization) can ``memorize'' all distinct pairs $\{z_t,x_{t+\tau}\}$ to get perfect reconstruction.  Why do we see non-perfect reconstruction for alanine dipeptide in practice?  I think it is due to two issues: 1. finite precision of floating numbers leads to ``fake finite state'', 2. neural network size is not large enough therefore two states with $z$ close to each other have to have similar reconstruction.  It may be easier to discuss these two issues in words, not in numerical estimation. 

Can we somehow transform the input features $(x, y)$ to equalize the variance explained? This operation would serve the same purpose as the whitening transformation in the linear case to eliminate the variance explained as a discriminating factor in the time-lagged reconstruction loss and force the TAE to identify the slow mode based on (generalized) autocorrelation alone. We first note that such a transform is not always possible even if we know the theoretical slow modes.  For instance, in our Washington beltway potential example, we cannot equalize $\sigma^2(\bar{x}(z\sim r))$ and $\sigma^2(\bar{x}(z\sim \theta))$ since the intrinsic circular symmetry assures that $\sigma^2(\bar{x}(z\sim r)) = 0$. What about other systems with slow modes that are possible to be theoretically whitened?  We note that for all but the most trivial systems we do not know the nonlinear slow modes \textit{a priori} and so there is no way to equalize the variance explained.  Even if we are able to identify putative slow modes using nonlinear dimensionality reduction techniques, if these modes are identified on the basis of anything other than slowness (e.g., variance) then equalization of the variance explained can still lead to incorrect results. If they are identified only on the basis of slowness then we have already obtained the slowest mode and there is no need to apply nonlinear TAE.  In short, there is no general procedure to correctly equalize the variance explained within the input features and therefore no way to guarantee that nonlinear TAEs will correctly discover the slowest mode.

\subsection{State-free reversible VAMPnets (SRVs) correctly identify nonlinear slow modes since the explained variance does not appear in the loss function}

\begin{figure}[ht!]
\begin{center}
	\includegraphics[width=0.50\textwidth]{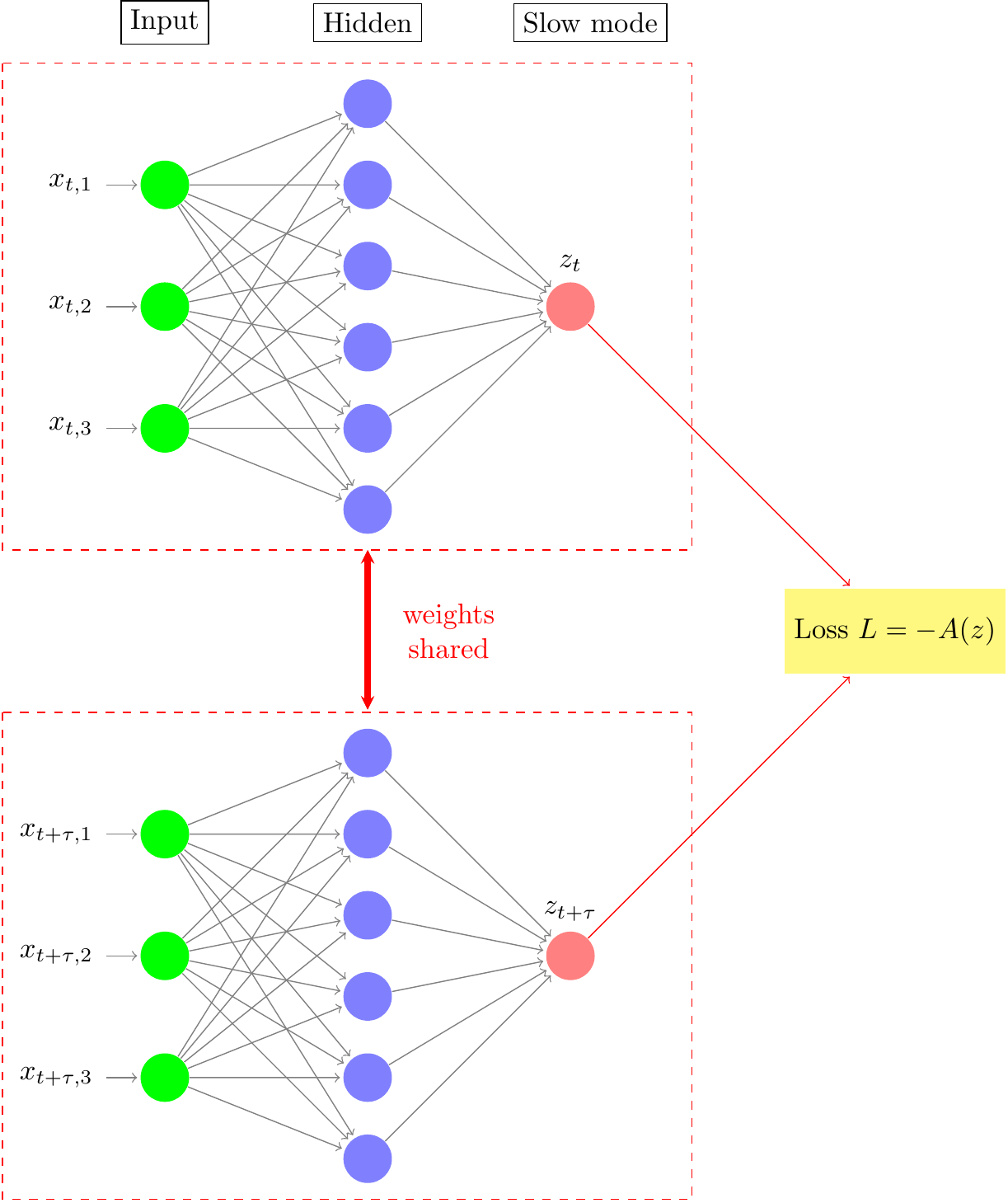}
	\caption{Schematic diagram of a 1D state-free reversible VAMPnet (SRV). A pair of input configurations $(x_t,x_{t+\tau})$ are fed into twin fully-connected feedforward neural network lobes with shared architectures and weights to generate network outputs $z_t$ and $z_{t+\tau}$. The neural network is trained with backpropagation to minimize the negative autocorrelation $L=-A(z)$.  Image constructed using code downloaded from \url{http://www.texample.net/tikz/examples/neural-network} with permission of the author Kjell Magne Fauske.}
	\label{SRV_diagram}
\end{center}
\end{figure}

We now demonstrate that in contrast to TAEs, state-free reversible VAMPnets (SRVs) can correctly identify the slowest mode using the same input features. Given a trajectory $\{x_t\}$, SRVs employ an artificial neural network to simultaneously discover the optimal nonlinear featurization of the input data $E(x_t) = \{E_j(x_t)\}_{j=1}^d$, where $E_j(x_t)$ is the $j^{\text{th}}$ component of the encoder output, and the linear combination of these features $z_i=\sum_{j=1}^d s_{ij}E_j(x_t)$ that maximizes the squared sum of autocorrelation of the slow modes. Full details of SRVs are presented in Ref.\ \cite{chen2019nonlinear}. A schematic of an SRV with $d$=1 corresponding to a 1D latent space embedding is presented in \blauw{Fig.~\ref{SRV_diagram}}. The corresponding loss function for the 1D SRV is,
\begin{equation}\label{SRV_loss}
d_{SRV}=-A(z)=-A(E(x_t)).
\end{equation}

We apply a SRV model with [2-50-50-1] architecture and tanh activation functions for all layers except input/output layers (an analogous design to the TAE above) to the whitened MSM trajectory on the Washington beltway potential. In \blauw{Fig.~\ref{C_TAE_CV}b}, we show the contour plot of the learned SRV mode that correctly identifies transitions in $r$ as the slowest mode of the system.  The reason why the SRV correctly identifies slowest mode is that the loss function is exactly equivalent to maximizing the autocorrelation of the learned mode with no contribution from variance explained.  Accordingly, we generally recommend to use a SRV rather than a TAE for estimation of the slowest mode.  Moreover, we note that when it comes to higher-order slow modes discovery through multidimensional latent variables, it is not recommended to use TAEs since there is no constraint that different components of the latent variable are orthogonal to each other, and it is therefore possible that each component becomes mixture of many slow modes.  In SRVs, however, the orthogonality constraints are satisfied naturally in the variational optimization procedure and it can discover a hierarchy of slow orthogonal modes \cite{chen2019nonlinear}.

\subsection{The ``slow'' mode learned by TAEs can be controlled by feature engineering}  

\begin{figure}[ht!]
	\begin{center}
		\includegraphics[width=\figurewidth]{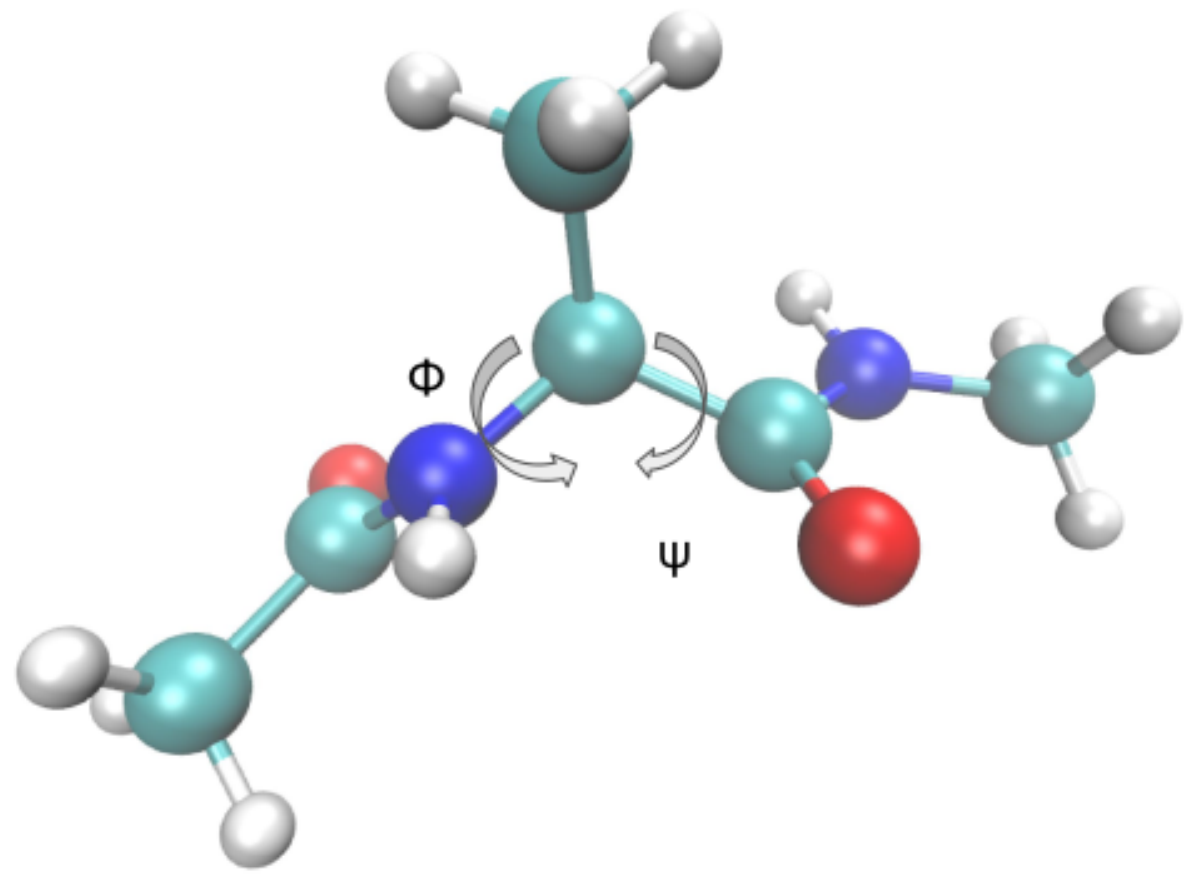}
			\caption{Molecular structure of alanine dipeptide with the two dominant backbone dihedral angles $\phi$ and $\psi$ annotated. Image rendered using VMD~\cite{humphrey1996vmd}.} \label{adp_molecule}
	\end{center}
\end{figure}

Following the ideas developed above, we now show that we can design features of a system to mislead nonlinear TAEs to learn a desired mode as the slowest mode simply by assuring that the target mode has much larger variance explained than the true slowest mode.  To show this, we employ molecular dynamics simulations of alanine dipeptide as 22-atom peptide that serves as the ``fruit fly'' for testing new numerical methods in molecular systems (\blauw{Fig.~\ref{adp_molecule}}).

\begin{figure*}[ht!]
	\begin{center}
		\includegraphics[width=0.65\textwidth]{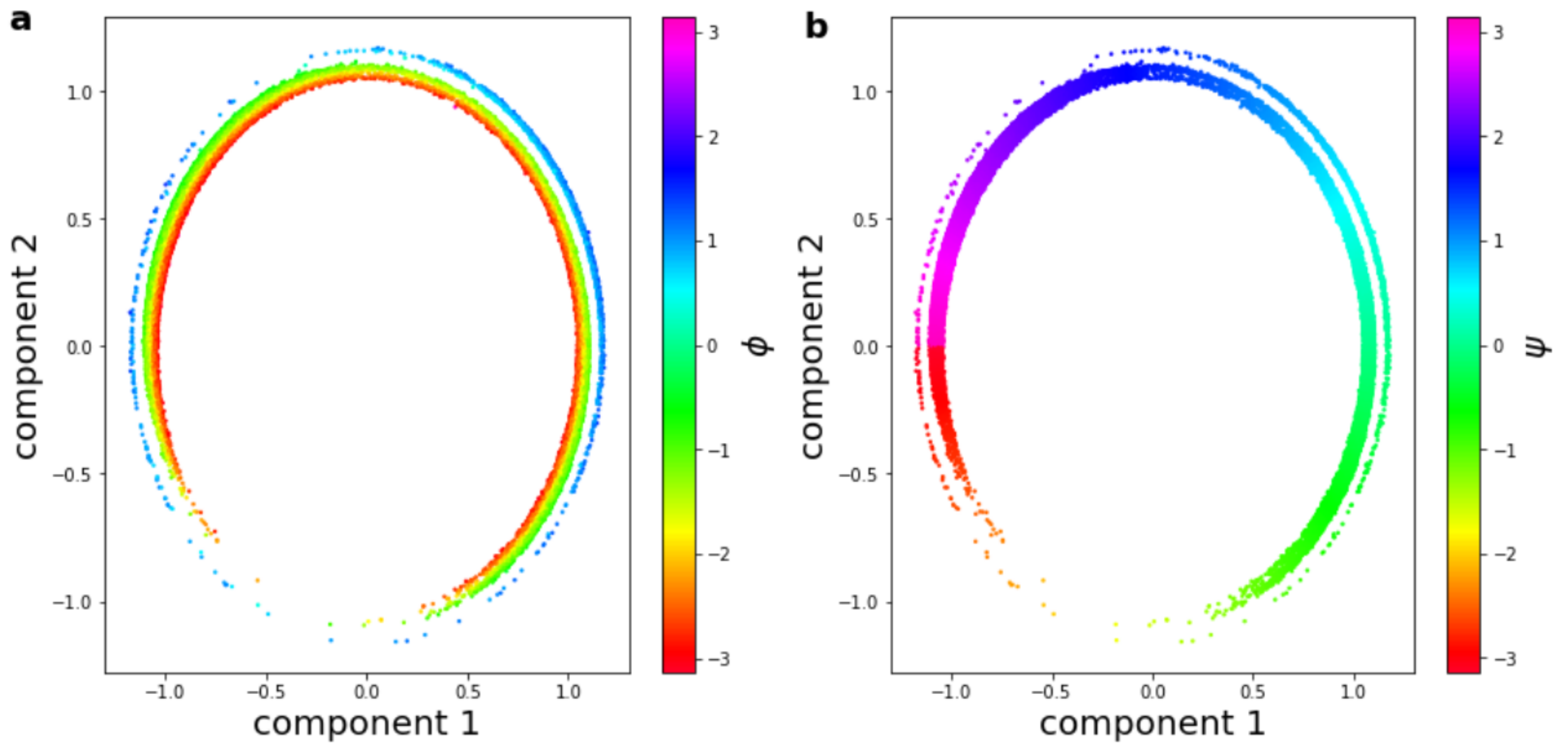}
		\caption{Input 2D features for alanine dipeptide transformed from two dihedral angles $\phi$ and $\psi$. (a) $\phi$ is encoded in radial direction while (b) $\psi$ is encoded in polar angle direction. }
		\label{feature_adp}
	\end{center}
\end{figure*}

It is well known from extensive prior study that the slowest mode for alanine dipeptide corresponds to transitions in the backbone dihedral angle $\phi$ \cite{chen2019nonlinear,trendelkamp2015estimation}. Is it possible to design a feature set such that a TAE is misled into learning the backbone dihedral angle $\psi$ as the slowest mode? To do so, consider 2D features $(x_1,x_2)$ given by,
\begin{align}\label{adp_features}
x_1=&r\cos \psi\nonumber\\
x_2=&r\sin \psi\nonumber\\
r =& r_0 + \Delta r \p{(\phi-2) \mod (2\pi) },
\end{align}
where $\Delta r << r_0$.  The idea is to map $(\phi,\psi)$ to a ring such that $\psi$ encodes polar angle direction while $\phi$ encodes radial direction such that $\psi$ has much larger variance explained than $\phi$.  Scatter plots of the engineered feature space colored by two dihedral angles $\phi$ and $\psi$ are presented in \blauw{Fig.~\ref{feature_adp}}.  

\begin{figure*}[ht!]
	\begin{center}
		\includegraphics[width=0.75\textwidth]{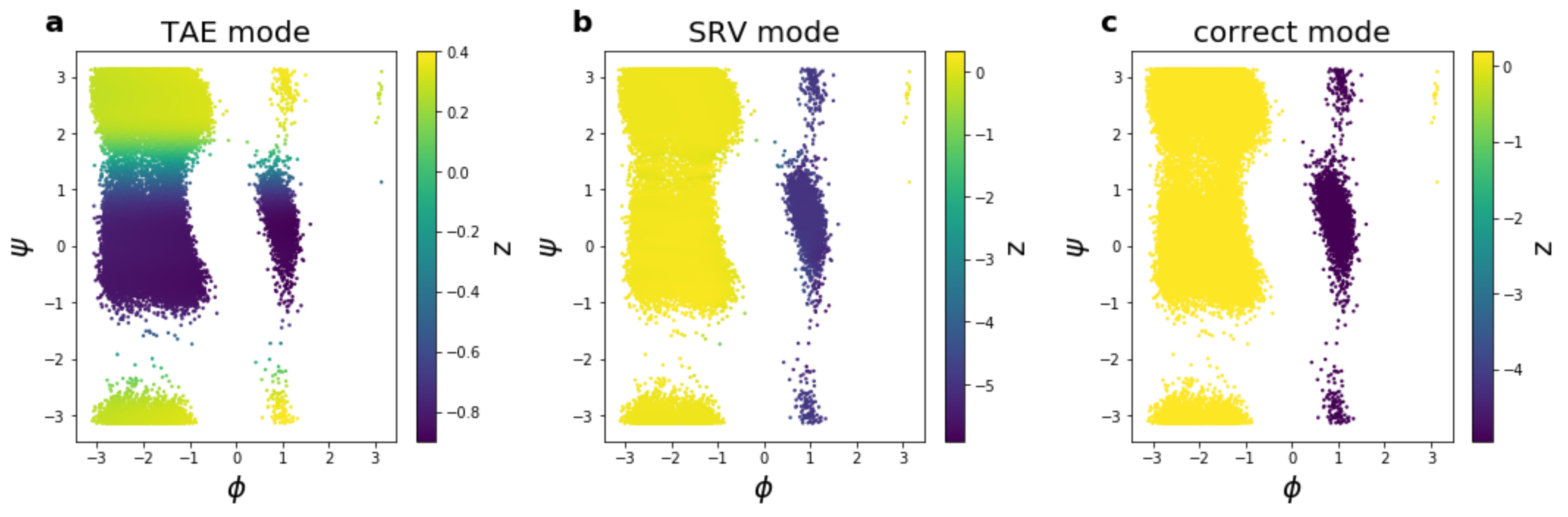}
		\caption{Slowest modes discovered by (a) TAE, (b) SRV with the features given by \blauw{Eq.~\ref{adp_features}} and (c) the ground truth slowest mode learned by a state-of-the-art MSM.  Misleading feature engineering causes the TAE to fail to discover the slowest mode whereas the SRV does so correctly using the same input features.}
		\label{adp_TAE_CV}
	\end{center}
\end{figure*}

We run molecular dynamics simulation for alanine dipeptide in explicit solvent to generate a trajectory of 2,000 ns saving frames every 2 ps to generate a 1,000,000-frame trajectory.  Then we use a TAE with [2-50-50-1-50-50-2] architecture and tanh activation functions for encoding and decoding layers and linear activation functions for input/output/latent layers to learn over the mean-free whitened trajectory data with features described above.   We show the Ramachandran plot colored by the slowest mode discovered by TAE in \blauw{Fig.~\ref{adp_TAE_CV}a}, which verifies that we successfully misled the TAE to learn $\psi$ as the slowest mode.  Again, the SRV employing an analogous architecture has no difficulty in correctly identifying the slowest mode (\blauw{Fig.~\ref{adp_TAE_CV}b}), which is consistent with the ground truth results of a state-of-the-art Markov state model trained in Ref.\ \cite{chen2019nonlinear} (\blauw{Fig.~\ref{adp_TAE_CV}c}).  

\begin{figure*}[ht!]
	\begin{center}
		\includegraphics[width=0.75\textwidth]{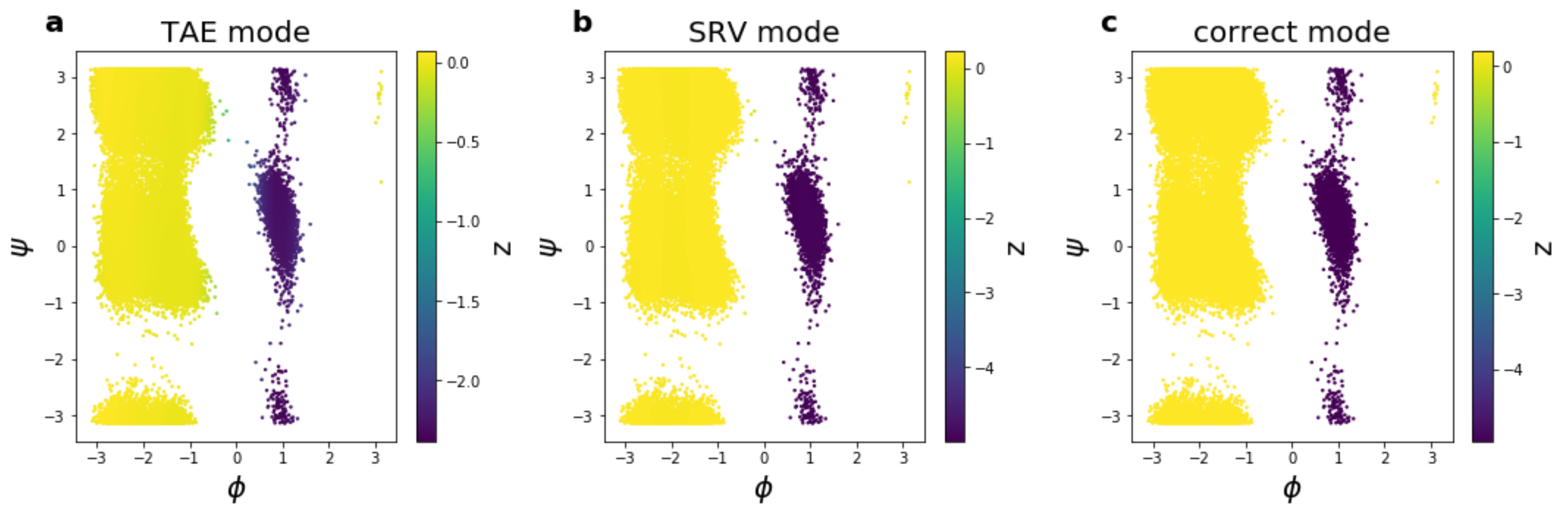}
		\caption{Slowest modes discovered by (a) TAE, (b) SRV with the features given by \blauw{Eq.~\ref{adp_features_1}} and (c) the ground truth slowest mode learned by a state-of-the-art MSM.  In this case, favorable feature engineering allows the TAE to correctly learn $\phi$ as slowest mode. Again, the SRV again correctly recovers the slowest mode from the same input features.}
		\label{adp_TAE_CV_1}
	\end{center}
\end{figure*}

Similarly if we swap $\phi$ and $\psi$ and define our features according to,
 \begin{align}\label{adp_features_1}
x_1=&r\cos \phi\nonumber\\
x_2=&r\sin \phi\nonumber\\
r =& r_0 + \Delta r \p{(\psi+2) \mod (2\pi) },
\end{align}
then the TAE learns $\phi$ as the slowest mode (\blauw{Fig.~\ref{adp_TAE_CV_1}a}), which is closer to the ground truth MSM results (\blauw{Fig.~\ref{adp_TAE_CV_1}c}), and again the SRV has no difficulty identifying slowest mode with these features (\blauw{Fig.~\ref{adp_TAE_CV_1}b}) 

\subsection{The 1D TAE structure can be modified to explicitly equalize the variance explained within the latent space to render it equivalent to a 1D SRV}  

To resolve the variance explained issue for nonlinear TAE, we can modify its structure. Instead of employing an autoencoding architecture to minimize the time-lagged reconstruction loss, we can instead employ an encoder-only architecture and optimize the time-lagged loss for the ``whitened'' encoder output.  This modification in architecture replaces the standard TAE loss given by \blauw{Eq.~\ref{loss}} with the modified TAE loss given by,
 \begin{equation}\label{modified_tae_loss}
d_\tau^M=\frac{\Expect{\norm{E(x_t)-E(x_{t+\tau})}^2}}{\sigma^2(E(x_t))},
\end{equation}
where $E(x_t)$ is the 1D encoder output for input $x_t$ at time $t$, the variance $\sigma^2(E(x_t))$ in the denominator is used to correctly ``whiten'' the learned slow mode such that the variance (or ``variance explained'') does not play a role in learning of the slowest mode.

How does the modified TAE architecture and loss function relate to the 1D SRV?  Following a similar derivation to \blauw{Eq.~\ref{tae_loss_linear}}, we have,
\begin{align}\label{key9}
&\Expect{\norm{E(x_t)-E(x_{t+\tau})}^2}\nonumber\\
%=&\Expect{\norm{\delta E(x_t)-\delta E(x_{t+\tau})}^2}\nonumber\\
=&2\sigma^2(E(x_t))(1-A(E(x_t))),
\end{align}
where $A(E(x_t))$ is the autocorrelation for $E(x_t)$.  Therefore \blauw{Eq.~\ref{modified_tae_loss}} becomes 
\begin{equation}\label{modified_tae_loss_2}
d_\tau^M=2-2A(E(x_t))
\end{equation}
which, up to a trivial affine transformation, is equivalent to the SRV loss given by \blauw{Eq.~\ref{SRV_loss}}. Accordingly, the  modified nonlinear TAE with loss given by \blauw{Eq.~\ref{modified_tae_loss_2}} is equivalent to a 1D SRV and can correctly identify slow modes without the misleading influence of variance explained.

\subsection{Variational dynamics encoders (VDEs) learn mixtures of the slowest mode and the maximum variance mode}  

Variational dynamics encoders (VDEs) employ a variational autoencoder architecture with a loss function comprising both reconstruction loss for inputs/outputs $x$ and the autocorrelation loss for the 1D latent variable $z$. The loss function of VDE can be written as~\cite{hernandez2018variational},
\begin{equation}\label{VDE_loss}
d_{VDE}=\lambda\p{\Expect{\norm{D(z_t)-x_{t+\tau}}^2}+L_{KL}}-(1-\lambda)A(z),
\end{equation}
where $D(z_t)$ is the reconstructed output, $\Expect{\norm{D(z_t)-x_{t+\tau}}^2}$ is the time-lagged reconstruction loss, $(-A(z))$ is the autocorrelation loss for $z$, and $\lambda$ is a linear mixing parameter.  If we ignore the Kullback-Leibler divergence term $L_{KL}$, which measures the similarity of the encoded probability distribution of $z$ to a Gaussian distribution, can be considered a form of regularization, and goes to zero in the case of well-trained variational autoencoders \cite{doersch2016tutorial}, then the loss is exactly the mixture of 1D SRV loss and 1D TAE loss.  Since SRVs learn the slowest mode, and TAEs learn a mixture of the slowest mode and the maximum variance mode, in general VDEs also learn a mixture of the slowest mode and the maximum variance mode. As was the case for TAEs, in any applications where we aim to find the slow modes it is not recommended to include terms associated with the explained variance (here within the reconstruction loss) and it is therefore not recommended to use VDEs for this goal.

\section*{Methods}

The TAE and SRV neural networks were constructed in Python using the Keras \cite{chollet2015keras} deep learning libraries and training performed on an NVIDIA GeForce GTX 1080 GPU card. The MSM simulations of particle motion over the ``Washington beltway'' potential were conducted in Python. Simulations of alanine dipeptide in water were conducted using the OpenMM 7.3 simulation suite \cite{eastman2012openmm,eastman2017openmm} employing the Amber99sb-ILDN forcefield for the biomolecule \cite{lindorff2010improved} and TIP3P forcefield for water \cite{jorgensen1983comparison}. The temperature and pressure were maintained at $T$ = 300 K and $P$ = 1 bar using Andersen thermostat \cite{andersen1980molecular} and $P$ = 1 atm using a Monte-Carlo barostat\cite{chow1995isothermal,aaqvist2004molecular}. Lennard-Jones interactions were smoothly switched to zero at a cuttoff of 1.4 nm cutoff, and Coulombic interactions were treated by particle-mesh Ewald \cite{EssmannPedersen1995} with a real space cutoff of 1.4 nm and a reciprocal space grid spacing of 0.12 nm.

\section*{Conclusions}

In this work, we present a theoretical analysis of the capabilities and limitations of TAEs in slow mode discovery.  We show that linear TAEs correctly learn the slowest linear mode with whitened features, while nonlinear TAEs cannot be assured of discovering the slowest nonlinear mode since it is not in general possible to perform an equivalent nonlinear ``whitening'' of the input features to equalize their variance explained.   We prove the theoretical bounds for time-lagged reconstruction loss for nonlinear TAE and demonstrate that a faster nonlinear mode could be erroneously identified as the slowest mode if it has significantly higher variance explained.  We validate our theoretical analysis in applications to a 2D ``Washington beltway'' potential and in molecular simulations of alanine dipeptide.  We also show how the 1D nonlinear TAE network structure can be modified to become equivalent to the 1D SRV to remove the misleading influence of variance explained and permit variable discrimination exclusively on the basis of autocorrelation. We also show that 1D VDEs mix the loss functions of TAEs and SRVs and therefore also suffer from the misleading influence of variance explained. Accordingly, SRVs serve as a more appropriate tool than TAEs or VDEs for the discovery of slow modes.  If variance explained of the modes is also of interest, then TAEs or VDEs can prove useful.

\section*{Acknowledgments}

This material is based upon work supported by the National Science Foundation under Grant No.~CHE-1841805.
H.S. acknowledges support from the Molecular Software Sciences Institute (MolSSI) Software Fellows program 
(NSF grant ACI-1547580)\cite{krylov2018perspective,wilkins2018nsf}.

\clearpage
\newpage

\onecolumngrid

\bibliography{library}

\clearpage
\newpage

\section*{\sffamily \Large Appendix}\label{appendix}

\setcounter{figure}{0}
\setcounter{section}{0}
\setcounter{subsection}{0}
\renewcommand{\thefigure}{A\arabic{figure}}

\subsection{Theoretical bounds for time-lagged autoencoder reconstruction loss}\label{theory_bounds}

We present a derivation of the theoretical bounds on the time-lagged autoencoder (TAE) reconstruction loss for a stationary process with finite states. We show that this leads to the approximate expression \blauw{Eq.~\ref{lower_bound_1}} in the main text, and that the approximation can be made arbitrarily accurate by employing sufficiently large neural network architectures. Let $x_t$ be the mean-free featurization of the system at time $t$ in a finite trajectory that therefore comprises a finite number of states.  The process is stationary and the trajectory is long enough such that for any $\tau\geq 0$,
\begin{align}\label{stationary}
\Expect{x_t}&=\Expect{x_{t+\tau}}=0,\nonumber\\
\Expect{x_t^2}&=\Expect{x_{t+\tau}^2}=\sigma^2(x),\nonumber\\
\end{align}
where $\sigma^2(x)$ is the total variance of featurized configurations.

Let $z_t=z(x_t)$ be the encoded embedding of $x_t$, where $z(\cdot)$ represents the encoding mapping. Let $f_t=f(z_t)$ be the decoded output of $z_t$, where $f(\cdot)$ represents the decoding mapping.  The time-lagged reconstruction loss $d_\tau(z)$ can be written as,
\begin{equation}\label{tae_loss}
d_\tau(z)=\Expect{\norm{x_{t+\tau}-f(z_t)}^2}.
\end{equation}

We define the expected evolution $\tilde{D}_\tau(z)$ after lag time $\tau$ given latent space encoding $z$ as,
\begin{equation}\label{tilde_D}
\tilde{D}_\tau(z)=\Expectsub{x_{t+\tau}}{x_{t+\tau}|z_t=z},
\end{equation}
which can be viewed as the ``optimal reconstructed output'' given $z$. Note that when $\tau=0$, $\bar{x}(z)=\tilde{D}_0(z)$ denotes the average system featurization corresponding to a particular latent-space encoding $z$.  

To quantify how much variance information of $x$ is captured by $z$, we define the variance explained,
\begin{equation}\label{variance_explained}
\sigma^2(\bar{x}(z))=\Expectsub{z}{\tilde{D}_0(z)^2},
\end{equation}
which measures the variance of the feature average given the encoding $z$ and is consistent with our definition \blauw{Eq.~\ref{ve}} in the main text. 

Now \blauw{Eq.~\ref{tae_loss}} becomes,
\begin{align}\label{tae_loss_3}
d_\tau(z)&=\Expect{\norm{x_{t+\tau}-f(z_t)}^2}\nonumber\\
&=\Expect{\norm{x_{t+\tau}-\tilde{D}_\tau(z_t)+\tilde{D}_\tau(z_t)-f(z_t)}^2}\nonumber\\
&=\Expect{\norm{x_{t+\tau}-\tilde{D}_\tau(z_t)}^2+\norm{\tilde{D}_\tau(z_t)-f(z_t)}^2 + 2 \p{x_{t+\tau}-\tilde{D}_\tau(z_t)}\p{\tilde{D}_\tau(z_t)-f(z_t)}}.\nonumber\\
\end{align}

Since we have finite number of possible $x_t$ values, the number of $z_t$ values should also be finite.  Therefore we can split the summation over the N-frame trajectory into two parts (sum-splitting trick): first summing over all frames with the same $z$ value, then summing over all $z$ values.  Considering the third term in \blauw{Eq.~\ref{tae_loss_3}},
\begin{align}\label{cross_term}
&\Expect{\p{x_{t+\tau}-\tilde{D}_\tau(z_t)}\p{\tilde{D}_\tau(z_t)-f(z_t)}}\nonumber\\
=&\frac{1}{N}\sum_t\p{x_{t+\tau}-\tilde{D}_\tau(z_t)}\p{\tilde{D}_\tau(z_t)-f(z_t)}\nonumber\\
=&\frac{1}{N}\sum_z\sum_{z_t=z}\p{x_{t+\tau}-\tilde{D}_\tau(z_t)}\p{\tilde{D}_\tau(z_t)-f(z_t)}\nonumber\\
=&\frac{1}{N}\sum_z\p{\tilde{D}_\tau(z)-f(z)}\sum_{z_t=z}\p{x_{t+\tau}-\tilde{D}_\tau(z_t)}\nonumber\\
\myxlongequal{Eq.~\ref{tilde_D}}&\frac{1}{N}\sum_z\p{\tilde{D}_\tau(z)-f(z)}\times 0=0.\nonumber\\
\end{align}

Therefore \blauw{Eq.~\ref{tae_loss_3}} becomes,
\begin{align}\label{tae_loss_4}
d_\tau(z)=&\Expect{\norm{x_{t+\tau}-\tilde{D}_\tau(z_t)}^2+\norm{\tilde{D}_\tau(z_t)-f(z_t)}^2 } \nonumber\\
=&\Expect{x_{t+\tau}^2}+\Expectsub{z}{\tilde{D}_\tau(z)^2}-2\Expect{x_{t+\tau}\tilde{D}_\tau(z_t)}+\Expect{\norm{\tilde{D}_\tau(z_t)-f(z_t)}^2 } \nonumber\\
\myxlongequal{Eq.~\ref{stationary}}&\sigma^2(x)+\Expectsub{z}{\tilde{D}_\tau(z)^2}-2\Expect{x_{t+\tau}\tilde{D}_\tau(z_t)}+\Expect{\norm{\tilde{D}_\tau(z_t)-f(z_t)}^2 }.\nonumber\\
\end{align}

We can apply the same sum-splitting idea to the third term of \blauw{Eq.~\ref{tae_loss_4}}, where $N$ is total number of frames, and $N(z)$ is the number of frames with encoding equal to $z$,
\begin{align}\label{cross_term_2}
&\Expect{x_{t+\tau}\tilde{D}_\tau(z_t)}\nonumber\\
=&\frac{1}{N}\sum_t x_{t+\tau}\tilde{D}_\tau(z_t)\nonumber\\
=&\frac{1}{N}\sum_z\tilde{D}_\tau(z)\sum_{z_t=z}x_{t+\tau}\nonumber\\
=&\frac{1}{N}\sum_z N(z)\tilde{D}_\tau(z)\p{\frac{1}{N(z)}\sum_{z_t=z}x_{t+\tau}}\nonumber\\
\myxlongequal{Eq.~\ref{tilde_D}}&\frac{1}{N}\sum_z N(z)\tilde{D}_\tau(z)\tilde{D}_\tau(z)\nonumber\\
=&\Expectsub{z}{\tilde{D}_\tau(z)^2}.
\end{align}

Therefore \blauw{Eq.~\ref{tae_loss_4}} becomes,
\begin{align}\label{tae_loss_6}
d_\tau(z)=&\sigma^2(x)-\Expectsub{z}{\tilde{D}_\tau(z)^2}+\Expect{\norm{\tilde{D}_\tau(z_t)-f(z_t)}^2 } \nonumber\\
=&\sigma^2(x)-\Expectsub{z}{\tilde{D}_0(z)^2}+\p{\Expectsub{z}{\tilde{D}_0(z)^2}-\Expectsub{z}{\tilde{D}_\tau(z)^2}}+\Expect{\norm{\tilde{D}_\tau(z_t)-f(z_t)}^2 } \nonumber\\
\myxlongequal{\blauw{Eq.~\ref{variance_explained}}}&\sigma^2(x)-\sigma^2(\bar{x}(z))+\Expect{\norm{\tilde{D}_\tau(z_t)-f(z_t)}^2 }+\p{\sigma^2(\bar{x}(z))-\Expectsub{z}{\tilde{D}_\tau(z)^2}}. \nonumber\\
\end{align}

If we define the ``generalized autocorrelation'' $G(z)$ as,
\begin{equation}\label{general_autocorr}
G(z)=\sqrt{1-\frac{\min_f \Expect{\norm{\tilde{D}_\tau(z_t)-f(z_t)}^2 }+\p{\sigma^2(\bar{x}(z))-\Expectsub{z}{\tilde{D}_\tau(z)^2}}}{\sigma^2(\bar{x}(z))}},
\end{equation}
then the lower bound of the TAE reconstruction loss is given by,
\begin{equation}\label{lower_bound}
d_\tau(z)\geq \sigma^2(\bar{x}(z))(1-G^2(z))+\p{\sigma^2(x)-\sigma^2(\bar{x}(z))}.
\end{equation}

Now we consider how good our lower bound is.  Due to the universal approximation theorem \cite{hassoun1995fundamentals, chen1995universal}, for any $\epsilon>0$, there exists a finite size decoder neural network $f$ such that the following inequality holds,
\begin{equation}\label{uat}
\norm{f(z_t)-\tilde{D}_\tau(z_t)}^2< \epsilon
\end{equation}
which means $f$ can be made arbitrarily close to the ``optimal reconstructed output'' as given in \blauw{Eq.~\ref{tilde_D}} by  employing a sufficiently large neural network.

Therefore \blauw{Eq.~\ref{tae_loss_6}} becomes
\begin{align}\label{tae_loss_7}
d_\tau(z)=&\sigma^2(x)-\sigma^2(\bar{x}(z))+\Expect{\norm{\tilde{D}_\tau(z_t)-f(z_t)}^2 }+\p{\sigma^2(\bar{x}(z))-\Expectsub{z}{\tilde{D}_\tau(z)^2}} \nonumber\\
<&\sigma^2(\bar{x}(z))(1-G^2(z))+\p{\sigma^2(x)-\sigma^2(\bar{x}(z))}+\epsilon
\end{align}

This indicates that the lower bound is actually quite tight, and it is a relatively good approximation of the TAE loss, yielding,
\begin{equation}
d_\tau(z)\approx  \sigma^2(\bar{x}(z))(1-G^2(z))+\p{\sigma^2(x)-\sigma^2(\bar{x}(z))},
\end{equation}
corresponding to \blauw{Eq.~\ref{lower_bound_1}} in the  main text.

What are the optimal encoding $z$ and the corresponding minimal possible loss for the TAE?  From \blauw{Eq.~\ref{tae_loss_6}} we see that if $f(z_t)=\tilde{D}_\tau(z_t)$, the optimal encoding $z$ should maximize $\Expectsub{z}{\tilde{D}_\tau(z)^2}$.  We define a finite set $S^z(z')=\{x|z(x)=z'\}$ that includes all configurations mapped to the same encoding value $z'$ under encoding mapping $z(\cdot)$.  If $z^{(1)}$ is a bijective encoding of $x$, then $\tilde{D}_\tau(z^{(1)})$ is,
\begin{equation}\label{key10}
\tilde{D}_\tau(z^{(1)})=\frac{1}{N(x(z^{(1)}))}\sum_{x_t=x(z^{(1)})}x_{t+\tau},
\end{equation}
where $N(x(z^{(1)}))$ is the number of frames with configuration $x_t=x(z^{(1)})$ and $x(z^{(1)})$ is the configuration corresponding to $z^{(1)}$.  For any encoding $z$, we have,
\begin{align}\label{key11}
\tilde{D}_\tau(z)^2=&\p{\frac{1}{N(z)}\sum_{z_t=z}x_{t+\tau}}^2\nonumber\\
=&\p{\frac{1}{N(z)}\sum_{x\in S^z(z)}\sum_{x_t=x}x_{t+\tau}}^2\nonumber\\
=&\p{\sum_{x\in S^z(z)}\frac{N(x)}{N(z)}\p{\frac{1}{N(x)}\sum_{x_t=x}x_{t+\tau}}}^2\nonumber\\
\leq & \sum_{x\in S^z(z)}\frac{N(x)}{N(z)}\p{\frac{1}{N(x)}\sum_{x_t=x}x_{t+\tau}}^2\nonumber\\
= & \sum_{x\in S^z(z)}\frac{N(x)}{N(z)}\tilde{D}_\tau(z^{(1)}(x))^2,\nonumber\\
\end{align}
where in the fourth line, we use Jensen's inequality, considering that,
\begin{equation}\label{key12}
\sum_{x\in S^z(z)}\frac{N(x)}{N(z)}=1.
\end{equation}
Therefore,
\begin{align}\label{key13}
\Expectsub{z}{\tilde{D}_\tau(z)^2}=&\frac{1}{N}\sum_z N(z)\tilde{D}_\tau(z)^2\nonumber\\
\leq & \frac{1}{N}\sum_z\sum_{x\in S^z(z)}N(x)\tilde{D}_\tau(z^{(1)}(x))^2\nonumber\\
=& \Expectsub{z}{\tilde{D}_\tau(z^{(1)})^2}.
\end{align}
This means that the optimal encoding that minimizes time-lagged reconstruction loss should be a bijective encoding of configurations.

\subsection{Interpretation of $G(z)$ as generalized autocorrelation}

We now demonstrate why $G(z)$ can be interpreted as a nonlinear generalization of the linear autocorrelation $A(z)$ that appears in the loss function for linear TAEs (\blauw{Eq.~\ref{slow_mode_def}}).  Consider a linear TAE applied on trajectory $\{x_t\}$ with independent components.   If $z_t$ is a linear transformation of the $k^\text{th}$ component of input features $x_{t,k}$, then $f(z_t)$ is also a linear transformation of $x_{t,k}$.  Without loss of generality and considering the independence among different components and that $z_t$ only includes information of $k^\text{th}$ component, all components of optimal output $f$ except $k^\text{th}$ component should be equal to 0, therefore we can write the encoding and decoding mapping as,
\begin{align}\label{encode_decode}
z_t=&c_1x_{t,k}+c_0\nonumber\\
f(z_t)_s=&\delta_{s,k}z_t\text{, } \delta_{s,k}=\begin{cases}
1 \text{, if } s=k\\
0\text{, otherwise}
\end{cases}.
\end{align}

Similarly the $s^\text{th}$ component of $\bar{x}(z_t)$ and $\tilde{D}_\tau(z_t)$ should be given by,
\begin{align}\label{encodings}
\p{\bar{x}(z_t)}_s=&\delta_{s,k}x_{t,k},\nonumber\\
\p{\tilde{D}_\tau(z_t)}_s=&\delta_{s,k}\Expect{x_{t+\tau, k}|x_{t,k}}.\nonumber\\
\end{align}
So all terms in $\min_f \Expect{\norm{\tilde{D}_\tau(z_t)-f(z_t)}^2 }+\p{\sigma^2(\bar{x}(z))-\Expectsub{z}{\tilde{D}_\tau(z)^2}}$ of \blauw{Eq.~\ref{general_autocorr}} have non-zero values only in their $k^\text{th}$ component.   

Now \blauw{Eq.~\ref{general_autocorr}} becomes,
\begin{align}\label{general_autocorr_3}
G(z)=&\sqrt{1-\frac{\min_f \Expect{\norm{\tilde{D}_\tau(z_t)-f(z_t)}^2 }+\p{\sigma^2(\bar{x}(z))-\Expectsub{z}{\tilde{D}_\tau(z)^2}}}{\sigma^2(\bar{x}(z))}}\nonumber\\
=&\sqrt{1-\frac{\min_f \Expect{f(z_t)^2}-2\Expect{f(z_t)\tilde{D}_\tau(z_t)} +\sigma^2(\bar{x}(z))}{\sigma^2(\bar{x}(z))}}\nonumber\\
\myxlongequal{\text{only $k^\text{th}$ components are non-zero}}&\sqrt{1-\frac{\min_{c_1,c_0} \Expect{(c_1x_{t,k}+c_0)^2}-2\Expect{(c_1x_{t,k}+c_0)\Expect{x_{t+\tau, k}|x_{t,k}}} +\sigma^2(x_{t,k})}{\sigma^2(x_{t,k})}}\nonumber\\
\myxlongequal{\text{\blauw{Eq.~\ref{sum_split}}}}&\sqrt{1-\frac{\min_{c_1,c_0} \Expect{(c_1x_{t,k}+c_0)^2}-2\Expect{(c_1x_{t,k}+c_0)x_{t+\tau, k}} +\sigma^2(x_{t,k})}{\sigma^2(x_{t,k})}}\nonumber\\
\myxlongequal{\text{\blauw{Eq.~\ref{stationary}}}}&\sqrt{1-\frac{\min_{c_1,c_0} \Expect{c_1^2x_{t,k}^2+c_0^2-2c_1x_{t,k}x_{t+\tau,k}} +\sigma^2(x_{t,k})}{\sigma^2(x_{t,k})}},\nonumber\\
\end{align}
where in the fourth line we use sum-splitting trick to simplify the second term in the numerator of the fraction,
\begin{align}\label{sum_split}
&\Expect{(c_1x_{t,k}+c_0)\Expect{x_{t+\tau, k}|x_{t,k}}}\nonumber\\
=&\frac{1}{N}\sum_t{(c_1x_{t,k}+c_0)\Expect{x_{t+\tau, k}|x_{t,k}}}\nonumber\\
\myxlongequal{\text{sum-splitting trick}}&\frac{1}{N}\sum_{x_{t,k}}N(x_{t,k})\sum_{x_{t',k}=x_{t,k}}{(c_1x_{t',k}+c_0)\Expect{x_{t'+\tau, k}|x_{t',k}}}\nonumber\\
=&\frac{1}{N}\sum_{x_{t,k}}N(x_{t,k})\sum_{x_{t',k}=x_{t,k}}{(c_1x_{t',k}+c_0)\frac{1}{N(x_{t',k})}\sum_{x_{t'',k}=x_{t',k}}x_{t''+\tau, k}}\nonumber\\
=&\frac{1}{N}\sum_{x_{t,k}}\p{\sum_{x_{t',k}=x_{t,k}}{(c_1x_{t',k}+c_0)\sum_{x_{t'',k}=x_{t',k}}x_{t''+\tau, k}}}\nonumber\\
=&\frac{1}{N}\sum_{x_{t,k}}\p{\sum_{x_{t',k}=x_{t,k}}{(c_1x_{t',k}+c_0)x_{t'+\tau, k}}}\nonumber\\
\myxlongequal{\text{sum-splitting trick}}&\frac{1}{N}\sum_{t}{(c_1x_{t,k}+c_0)x_{t+\tau, k}}\nonumber\\
=&\Expect{(c_1x_{t,k}+c_0)x_{t+\tau, k}},
\end{align}
where $N(x_{t,k})$ is number of frames with $k^\text{th}$ component equal to $x_{t,k}$, $\sum_{x_{t,k}}$ denotes summation over all possible $x_{t,k}$ values, and $\sum_{x_{t',k}=x_{t,k}}$ denotes summation over all frames with $k^\text{th}$ component equal to $x_{t,k}$.

The optimal $c_0$ and $c_1$ follow from minimization with respect to these parameters and satisfy,
\begin{align}\label{c0c1}
c_0=&0,\nonumber\\
c_1=&\frac{\Expect{x_{t,k}x_{t+\tau,k}}}{\Expect{x_{t,k}^2}}=A(x_{t,k}),
\end{align}
where $A(x_{t,k})$ is the traditional autocorrelation of component $x_{t,k}$.

So \blauw{Eq.~\ref{general_autocorr_3}} becomes,
\begin{align}\label{general_autocorr_4}
G(z)=&\sqrt{1-\frac{\min_{c_1,c_0} \Expect{c_1^2x_{t,k}^2+c_0^2-2c_1x_{t,k}x_{t+\tau,k}} +\sigma^2(x_{t,k})}{\sigma^2(x_{t,k})}}\nonumber\\
=&\sqrt{1-\frac{A^2(x_{t,k})\Expect{x_{t,k}^2}-2A(x_{t,k})\Expect{x_{t,k}x_{t+\tau,k}} +\sigma^2(x_{t,k})}{\sigma^2(x_{t,k})}}\nonumber\\
=&\sqrt{-\frac{A^2(x_{t,k})\sigma^2(x_{t,k})-2A(x_{t,k})\sigma^2(x_{t,k})A(x_{t,k})}{\sigma^2(x_{t,k})}}\nonumber\\
=&A(x_{t,k})=A(z).
\end{align}
We see that in the linear case with independent components, the generalized autocorrelation $G(z)$ reduces to the standard linear autocorrelation $A(z)$.

In the nonlinear case, the minimal possible loss in \blauw{Eq.~\ref{tae_loss_6}} is obtained if $f(z_t)=\tilde{D}_\tau(z_t)$, wherein $G(z)$ becomes,
\begin{equation}\label{general_autocorr_5}
G(z)=\sqrt{\frac{\Expectsub{z}{\tilde{D}_\tau(z)^2}}{\sigma^2(\bar{x}(z))}}.
\end{equation}

We note the following two attractive properties of $G(z)$. First, $G(h(z))=G(z)$ if $h$ is a bijection. This is important for nonlinear TAEs since under the transformation $z\to h(z), f\to f\circ h^{-1}$ the TAE loss is invariant and $G(\cdot)$ is  invariant, whereas the autocorrelation $A(\cdot)$ is not invariant. Second, $G(z)$ can be naturally applied to multi-dimensional $z$ where it is difficult to define the traditional autocorrelation $A(z)$.

\end{document}